\documentclass[10pt,twocolumn,letterpaper]{article}

\usepackage[pagenumbers]{cvpr} 

\definecolor{cvprblue}{rgb}{0.21,0.49,0.74}
\usepackage[pagebackref,breaklinks,colorlinks,allcolors=cvprblue]{hyperref}
\usepackage{booktabs}
\usepackage{graphicx} 
\usepackage{array}
\usepackage{caption}
\usepackage{amsmath}
\usepackage{adjustbox}
\usepackage{xcolor}
\usepackage{multirow}

\def\paperID{405} 
\def\confName{CVPR}
\def\confYear{2025}

\title{Prompt-Guided Environmentally Consistent Adversarial Patch}

\author{
    Chaoqun Li\\
    Tsinghua University\\
    \and
    Huanqian Yan\\
    Tsinghua University\\
    \and
    Lifeng Zhou\\
    Anhui University\\
    \and
    Tairan Chen\\
    Tsinghua University\\
    \and
    Zhuodong Liu\\
    Tsinghua University\\
    \and
    Hang su\\
    Tsinghua University\\
}

\begin{document}
\maketitle
\begin{abstract}
Adversarial attacks in the physical world pose a significant threat to the security of vision-based systems, such as facial recognition and autonomous driving. Existing adversarial patch methods primarily focus on improving attack performance, but they often produce patches that are easily detectable by humans and struggle to achieve environmental consistency, i.e., blending patches into the environment. This paper introduces a novel approach for generating adversarial patches, which addresses both the visual naturalness and environmental consistency of the patches. We propose \textbf{Prompt-Guided Environmentally Consistent Adversarial Patch (PG-ECAP)}, a method that aligns the patch with the environment to ensure seamless integration into the environment. The approach leverages diffusion models to generate patches that are both environmental consistency and effective in evading detection. To further enhance the naturalness and consistency, we introduce two alignment losses: \textbf{Prompt Alignment Loss} and \textbf{Latent Space Alignment Loss}, ensuring that the generated patch maintains its adversarial properties while fitting naturally within its environment. Extensive experiments in both digital and physical domains demonstrate that PG-ECAP outperforms existing methods in attack success rate and environmental consistency. 

\end{abstract}

\begin{figure}[t]
  \centering
  \includegraphics[width=1.0\columnwidth]{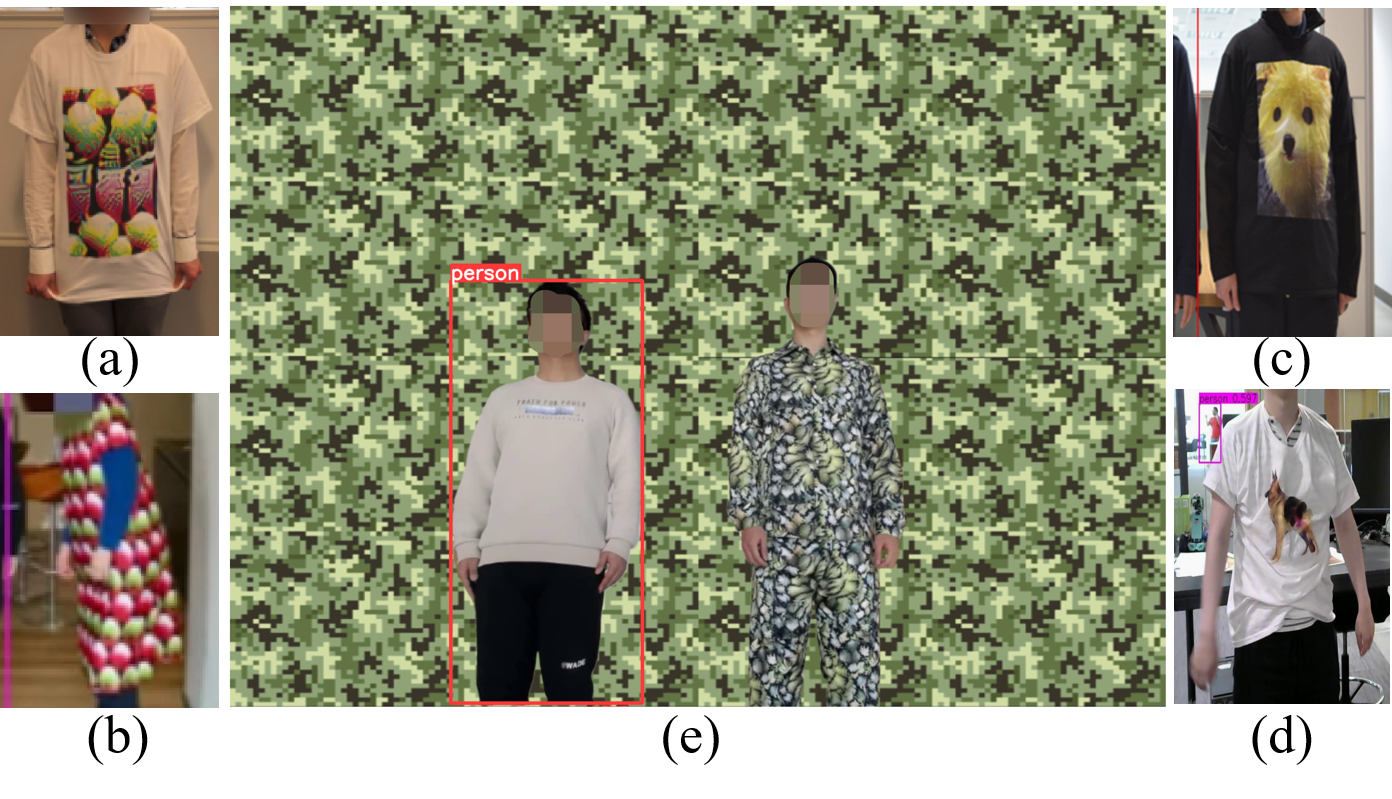}
  \caption{A comparison of various adversarial clothes: (a) Adversarial T-shirt~\cite{xu2020adversarial}, (b) Adversarial Texture~\cite{hu2022adversarial}, (c) NAP~\cite{hu2021naturalistic}, (d) DAP~\cite{guesmi2024dap}, and (e) Ours. We cover our generated adversarial patch onto the long-sleeved T-shirt to create our adversarial clothing. Among these methods, only our clothing consistent with the environment, making it more practical in real-world scenarios.
}
  \label{pics_first}
\end{figure}

\section{Introduction}
\label{sec:intro}
Adversarial attacks pose a significant threat to various fields, such as image classification~\cite{rawat2017deep}, object detection~\cite{zhao2019object}, speech recognition~\cite{qin2019imperceptible}, and natural language processing~\cite{zhang2020adversarial}. Generally, adversarial attacks can be categorized into digital attacks and physical attacks: digital attacks deceive models by modifying digital inputs, while physical attacks achieve this by modifying the real-world objects. Compared with digital attacks, physical attacks pose a greater threat in high-security, real-world applications such as facial recognition~\cite{dong2019efficient}, autonomous driving~\cite{cao2019adversarial}, and surveillance systems~\cite{thys2019fooling}. 

\begin{figure}[t]
  \centering
  \includegraphics[width=1.0\columnwidth]{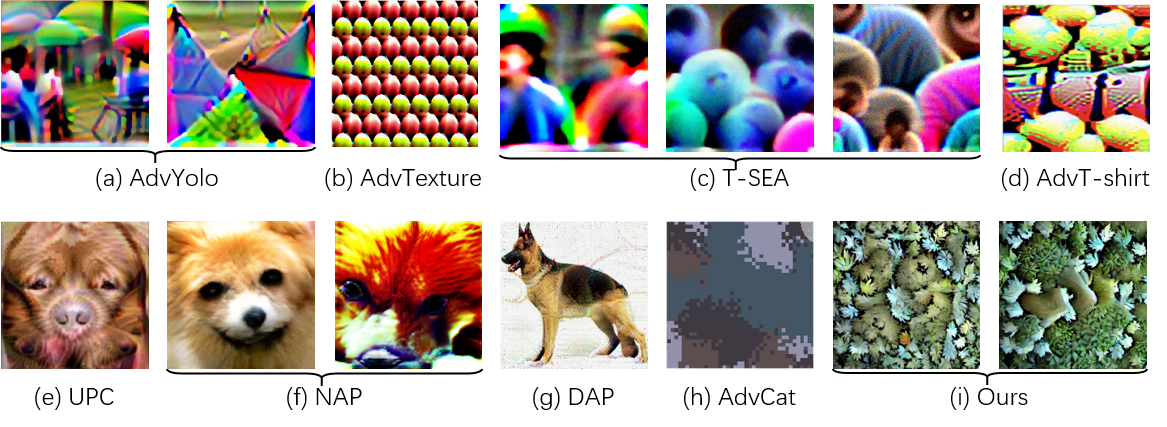}
  \caption{Ours vs State-of-the-Art patches: (a) AdvYolo~\cite{thys2019fooling}, (b) AdvTexture~\cite{hu2022adversarial}, (c) T-SEA~\cite{huang2023t}, (d) AdvT-shirt~\cite{xu2020adversarial}, (e) UPC~\cite{huang2020universal}, (f) NAP~\cite{hu2021naturalistic}, (g) DAP~\cite{guesmi2024dap}, (h) AdvCat~\cite{hu2023physically}, and (i) Ours. Our patches achieve a more natural and environmentally consistent appearance in forest-like environments.
  }
  \label{compare}
\end{figure}

Unlike digital attacks, which can precisely manipulate inputs in a fully controlled environment, physical attacks must handle with unpredictable real-world conditions. Factors such as lighting, angles, surface textures, and environmental noise can significantly degrade their effectiveness. Adversarial patches are a common form of physical attack, typically attached as specific patterns on objects’ surface to cause deep learning models misclassifying or ignoring the objects. Early methods for generating adversarial patches focus extremely on attack performance~\cite{xu2020adversarial, hu2022adversarial, huang2023t, wu2020making}, often resulting in colorful patches that are easily noticeable by humans (see Fig.~\ref{compare} (a), (b), (c), and (d)). Subsequent research focuses on the naturalness of patches, such as generating adversarial patches based on images of cats and dogs~\cite{hu2021naturalistic, guesmi2024dap, huang2020universal, wang2021dual}. However, these generated patches often exhibit noticeable distortions compared to reference images, making them easily observed (see Fig.~\ref{compare} (e), (f), and (g)). Additionally, these methods overlook the importance of making the patch blend naturally with the environment—known as \textbf{environmental consistency}—which reduces their practicality in real-world scenarios. A later approach focuses on generating patches with limited colors~\cite{hu2023physically}, but it relies heavily on fixed color spaces, making it difficult to maintain consistency with new environments. Moreover, sudden color changes in the patch reduce its natural appearance (see Fig.~\ref{compare} (h)).

Based on the limitations of existing methods, we aim to develop a new approach with two key properties. First, it should generate adversarial patches that are visually natural and effective in attack, addressing the issue of human-noticeable distortions present in prior methods. Second, our method should have the ability to ensure the environmental consistency of adversarial patches, generating patches that adapt to the environment.


To achieve this, we propose a novel method called \textbf{P}rompt-\textbf{G}uided \textbf{E}nvironmentally \textbf{C}onsistent \textbf{A}dversarial \textbf{P}atch (\textbf{PG-ECAP}). Totally, PG-ECAP uses the text-to-image ability of diffusion models~\cite{rombach2022high} and integrates two alignment losses to achieve our goal. Specifically, to ensure the visual naturalness of patches, we adopt diffusion models~\cite{rombach2022high} that are trained on a wide range of real images, enabling them to create visually natural images. Furthermore, to incorporate environmental constraints into adversarial patch generation, we use textual prompts to introduce these constraints. By providing an appropriate textual prompt that reflects the characteristics of a specific environment (e.g., green patterns for a forest-like environment or desert-themed patterns for a desert-like environment), we can effectively generate an adversarial patch that is consistent with that environment. When the environment changes, we can easily adjust the prompt to match the new environment, enabling us to generate adversarial patches that are adapted to the new environment (see Fig.~\ref{ablation_de}). 
In contrast, previous methods often rely on fixed color spaces.

Building on previous studies~\cite{chen2024diffusion, chen2024content}, we extend adversarial patch attacks into the latent space of diffusion models, ensuring that the generated patch enables objects to evade detection. However, the balance between naturalness and attack performance remains a concern. Without appropriate constraints, the generated patch may significantly differ from the prompt (see Fig.~\ref{distortion}). To address this, we introduce the Prompt Alignment Loss and the Latent Space Alignment Loss. The Prompt Alignment Loss preserves the cross attention maps between the prompt and latent variables in the diffusion model, enabling the model to generate an adversarial patch that is aligned with the prompt. Meanwhile, the Latent Space Alignment Loss preserves the initial latent variable and aligns it with the newly generated latent variable during optimization in the latent space, further enhancing the environmental consistency of the generated patch. As shown in Fig.\ref{pics_first}, we tile our generated patch onto a long-sleeved T-shirt. Compare to other methods, only our method can consistent with the environment.  

The main contributions of our work are as follows:
\begin{itemize}

\item We introduce PG-ECAP, an adversarial patch generation method that leverages the text-to-image capabilities of diffusion models~\cite{rombach2022high} to create natural and environmentally consistent patches.
\item We propose two novel alignment losses, Prompt Alignment Loss and Latent Space Alignment Loss, which ensure that the generated patches align with the environmental context from both the prompt and latent space perspectives.
\item Our extensive experimental results, conducted in both digital and physical domains, demonstrate that PG-ECAP outperforms existing methods in terms of environmental consistency and attack success rate, validating its effectiveness and practicality across various real-world scenarios.
\end{itemize}


\begin{figure*}[t]
  \centering
  \includegraphics[width=1.0\textwidth]{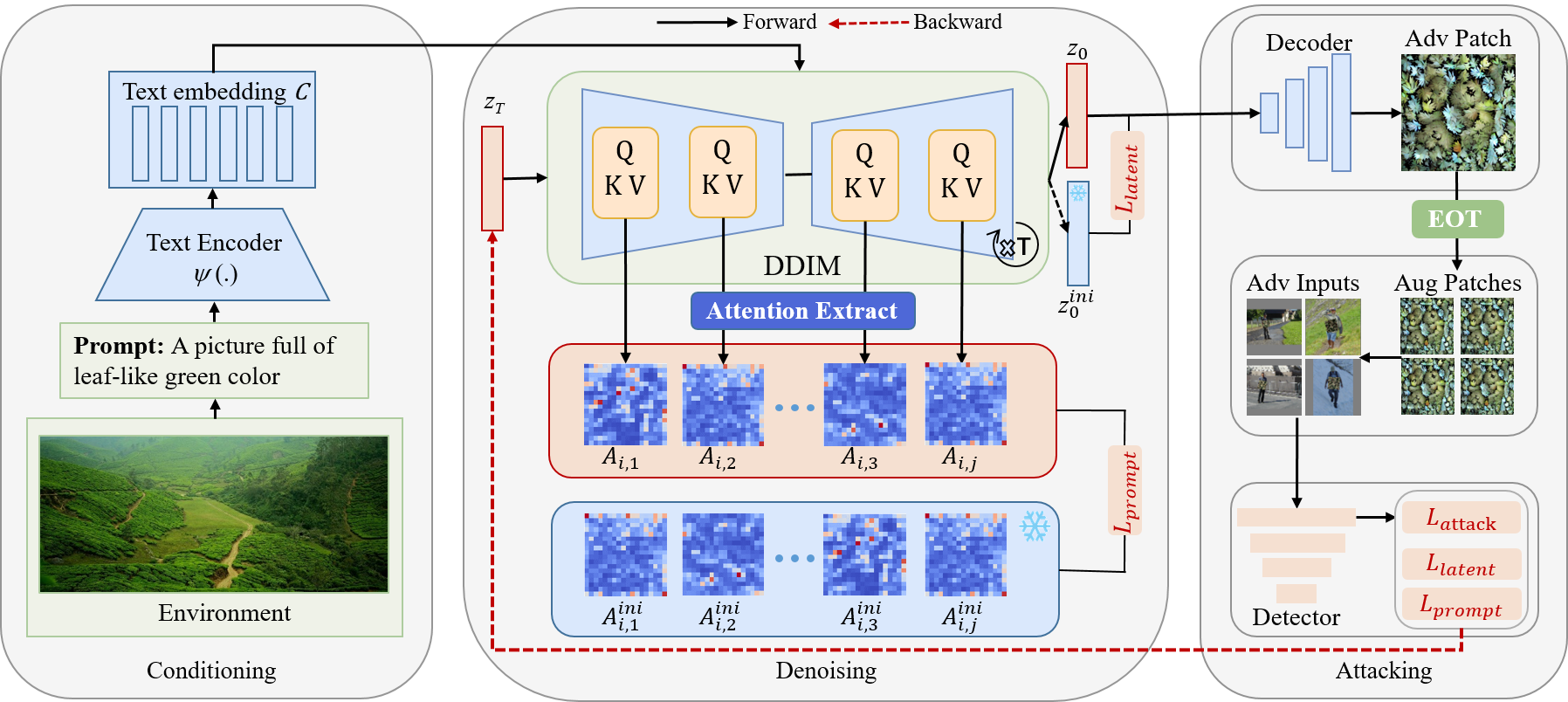}
  \caption{\textbf{An overview of the proposed PG-ECAP.} We first extract prompt $\mathcal{P}$ from the environment and feed $\mathcal{P}$ into a text encoder \( \psi(.) \) to obtain text embeddings $\mathcal{C}$. Then, we sample \( z_T \) from a Gaussian distribution and feed it with $\mathcal{C}$ into the diffusion model, extracting the cross attention maps during the DDIM process for alignment. After the DDIM process is finished, we align both the cross attention maps and \( z_0 \) with their corresponding initial values to ensure that the generated image aligns with $\mathcal{P}$. Finally, we decode the optimized \( z_0 \) to obtain the adversarial patch, augment it using EOT, and attach them onto the images to generate adversarial inputs. These inputs are then fed into the model to obtain detection confidence scores, which serve as the loss function to optimize \( z_T \) with the two alignment losses.
}
  \label{pipeline}
\end{figure*}

\section{Related Work}
In this section, we first provide an overview of diffusion models and their applications, followed by an introduction to physical adversarial attacks against object detectors.
\subsection{Diffusion Models}
\label{diffusion models}
Diffusion models have made remarkable progress in recent years. Early works, such as Denoising Diffusion Probabilistic Models (DDPM)~\cite{ho2020denoising}, lays the foundation and shows the potential for image synthesis. Building on this foundation, Denoising Diffusion Implicit Models (DDIM)~\cite{song2020denoising} are introduced as a non-Markovian extension of DDPM, enabling fewer diffusion steps while still preserving high-quality image synthesis. The introduction of Latent Diffusion Models (LDM) by Rombach et al.~\cite{rombach2022high} further enhances this approach, enabling more efficient image generation in the latent space. A significant milestone is achieved by Stable Diffusion~\cite{ramesh2021zero}, which integrates CLIP embeddings~\cite{radford2021learning} to improve the alignment between generated images and their corresponding prompts. This integration greatly enhances the relevance of the generated images to the prompts.

Subsequent research expands the ability of diffusion models in various directions. Liu et al.~\cite{liu2022text} introduce text-guided image editing, while Chen et al.~\cite{chen2023multimodal} develop multimodal diffusion models capable of handling both text and image data. To address the challenge of processing long text descriptions, Li et al.~\cite{li2023hierarchical} refine the image generation process step-by-step. Chen et al.~\cite{chen2024diffusion, chen2024content} further apply diffusion models to attack deep classifiers. Despite these numerous applications mentioned above, the potential of diffusion models for attacking object detectors in the physical world has not been fully explored.

\subsection{Physical Adversarial Attacks}
\label{physical attacks}
Since Thys et al.~\cite{thys2019fooling} propose adversarial patches against pedestrian detectors, this area attracts extensive research. Beyond pedestrians, Duan et al.~\cite{duan2021learning} extend the attack categories to vehicles. Wu et al.~\cite{wu2020making} and Xu et al.~\cite{xu2020adversarial} explore evading detectors by printing adversarial patches on clothing. Specifically, Xu et al.~\cite{xu2020adversarial} address the challenge of reduced attack performance by simulating non-rigid deformations of clothing. Furthermore, to achieve successful evasion from multiple perspectives in the physical world, several methods~\cite{duan2021learning, wang2022fca, hu2022adversarial} develop techniques to ensure full coverage of adversarial patches on objects, leading to 3D adversarial attacks. Among these, Duan et al.~\cite{duan2021learning} and Wang et al.~\cite{wang2022fca} use a neural renderer~\cite{kato2018neural} to render adversarial patches onto objects, while Hu et al.~\cite{hu2022adversarial} employ toroidal cropping to guarantee full coverage. Additionally, some works focus on enhancing the transferability of adversarial patches. For example, Huang et al.~\cite{huang2023t} mitigate the overfitting effect of adversarial patches on white-box models using a series of data augmentation techniques. However, the methods mentioned above primarily focus on attack performance, resulting in patches that are often overly conspicuous and easily noticeable by humans.

Subsequent works focus on improving the naturalness of adversarial patches~\cite{wang2021dual, hu2021naturalistic, huang2020universal, guesmi2024dap}. Specifically, ~\cite{huang2020universal, guesmi2024dap} preserve the naturalness of generated patches by modifying reference images (such as common animals like cats or dogs) in pixel space.~\cite{hu2021naturalistic} employs GANs~\cite{Brock2019biggan} to generate adversarial patches that look like specific animals.~\cite{wang2021dual} creates natural adversarial patches by suppressing both model and human attention. However, these methods often exhibit noticeable distortions compared to typical animal images and lack environmental consistency. A more recent method~\cite{hu2023physically} generates adversarial patches by selecting colors from a limited color space to create patterns that look like natural designs, such as camouflage. However, it heavily relies on fixed color spaces, limiting its adaptability to different environments. Moreover, the sudden color changes in the patch reduce its natural appearance.

Given the limitations of these methods, our goal is to develop a new approach that generates adversarial patches that are both visually natural and consistent with the environment. Additionally, the proposed method should be able to easily generate new patches adapted to new environments.

\section{Methodology}
\label{sec:methodology}
As illustrated in Fig.~\ref{pipeline}, PG-ECAP utilizes the text-to-image capabilities of diffusion models and incorporates two alignment losses to enhance the patches' consistency with their environment. 
We first introduce the text-to-image generation process based on diffusion models, followed by the problem formulation and our optimization strategy.

\subsection{Preliminary}
\label{preliminary}
In this section, we detail the text-to-image generation process to clarify the remain parts of our method. The goal of the process is to generate a sample \( x_0 \) conditioned on an input prompt \( \mathcal{P} \). 

First, the prompt \( \mathcal{P} \) is encoded by a text encoder \( \psi(.) \), resulting in text embeddings \( \mathcal{C} = \psi(\mathcal{P}) \) to guide latent variables in diffusion models. Then, a latent variable \( z_T \) is drawn from a standard normal distribution \( \mathcal{N}(0, \mathbf{I}) \). The next step is iteratively denoising \(z_T \) to obtain \(z_0 \) through sampling from the posterior Gaussian distribution \(q\left(z_{t-1} \mid z_t, \mathcal{C}\right) \). Due to the \( q\left(z_{t-1} \mid z_t, \mathcal{C}\right) \) is intractable, a diffusion model \( p_\theta \) is trained to approximate this posterior by predicting the mean \(\mu_\theta \) and covariance \(\Sigma_\theta \)~\cite{sohl2015deep}:
\begin{equation}
\centering
p_\theta\left(z_{t-1} \mid z_t, \mathcal{C}\right)=\mathcal{N}\left(\mu_\theta\left(z_t, t, \mathcal{C}\right), \Sigma_\theta\left(z_t, t, \mathcal{C}\right)\right)    
\end{equation}
The mean \(\mu_\theta \) is further expressed as:
\begin{equation}
\centering
\mu_\theta\left(z_t, t, \mathcal{C}\right)=\sqrt{\frac{1}{\alpha_t}}\left(z_t-\frac{\beta_t}{ \sqrt{1-\bar{\alpha}_t}} \epsilon_\theta\left(z_t, t, \mathcal{C}\right)\right)    
\end{equation}
where \(\alpha_t \) and \( \beta_t \) are noise schedule parameters, \(\bar{\alpha}_t \) is the cumulative product of \(\alpha_t \), and \(\epsilon_\theta\left(z_t, t, \mathcal{C}\right) \) is the predicted noise from the diffusion model.

This noise prediction integrates multiple operations, including a cross attention mechanism, enabling the model to effectively leverage information from the $\mathcal{C}$. The mathematical representation of the cross attention can be expressed as:
\begin{equation}
\centering
\label{cross_attention}
\operatorname{Attention}(Q, K, V)=\operatorname{softmax}\left(\frac{Q K^T}{\sqrt{d_k}}\right) V
\end{equation}
where the query \( Q \) is based on features of latent variables, while the keys \( K \) and values \( V \) are based on text embeddings $\mathcal{C}$.

The optimization of the model \( p_\theta \) is guided by the following loss function, which quantifies the difference between the true noise \( \epsilon \) and the predicted noise~\cite{ho2020denoising}:
\begin{equation}
\centering
\min _\theta L(\theta)=\mathbb{E}_{x_0, \epsilon \sim \mathcal{N}(0, \mathbf{I}), t}\left\|\epsilon-\epsilon_\theta\left(z_t, t, \mathcal{C}\right)\right\|_2^2
\end{equation}

After training the model \( p_\theta \), the sampling process is defined as:
\begin{equation}
\centering
z_{t-1}=\mu_\theta\left(z_t, t, \mathcal{C}\right)+\sigma_t \epsilon, \quad \epsilon \sim \mathcal{N}(0, \mathbf{I})
\end{equation}
where \( \sigma_t \) represents the standard deviation controlling the noise.

To enhance the efficiency of this sampling process, Denoising Diffusion Implicit Models (DDIM)~\cite{song2020denoising} generalize the classical DDPMs by introducing a non-Markovian framework. In DDIM, the sampling process is modified to:

\begin{equation}
\centering
\begin{split}
z_{t-1} &= \sqrt{\bar{\alpha}_{t-1}} \left( \frac{z_t - \sqrt{1 - \bar{\alpha}_t} \epsilon_\theta(z_t, t, \mathcal{C})}{\sqrt{\bar{\alpha}_t}} \right) \\
& + \sqrt{1 - \bar{\alpha}_{t-1} - \sigma_t^2} \epsilon_\theta(z_t, t, \mathcal{C}) + \sigma_t \xi, \quad \xi \sim \mathcal{N}(0, \mathbf{I}).
\end{split}
\end{equation}
By setting \( \sigma_t = 0 \), DDIM achieves a deterministic sampling process that allows for improved efficiency in generating images.

Once the iterative process reaches time step \( t=0 \), we obtain the latent variable \( z_0 \), which contains rich semantic information from \( \mathcal{P} \) through cross attention mechanism. At this stage, \( z_0 \) is decoded into the final image \( x_0 \) using a decoder network that transforms the latent representation back into pixel space. The overall pipeline is formulated as:
\begin{equation}
\left\{
\begin{aligned}
z_T & \sim \mathcal{N}(0, \mathbf{I}) \\
z_0 & = \mu_\theta(\mu_\theta(\cdots,\mu_\theta(z_T, T, \mathcal{C}) + \sigma_T \epsilon_T,\cdots, 1, \mathcal{C}) + \sigma_1 \epsilon_1 \\
x_0 & = \text{Decoder}(z_0)
\end{aligned}
\right.
\label{diffusion_pipeline}
\end{equation}
where the second term of this equation represents the iterative process of denoising \( z_T \) to \(z_0 \).

\subsection{Problem Formulation}
\label{problem_formulation}

In adversarial attacks against object detectors, the objective is to design an adversarial patch \( \mathcal{Q} \) that causes the object detector \( M \) to fail in detecting objects within an image \( I \). Traditional adversarial patch methods aim to minimize detection accuracy, but they often overlook the critical aspect of environmental consistency—ensuring that the patch integrates naturally with its surroundings under real-world conditions. For an attack to be effective and natural, the patch must not only be imperceptible to the detector but also blend seamlessly with the environment, despite variations in lighting, viewpoint, texture, and other factors.

We define the adversarial patch generation problem as a joint optimization that minimizes detection confidence while enforcing environmental consistency. Specifically, we aim to generate a patch that is both effective at evading detection and visually harmonious with the background, given a set of environmental transformations \( \Phi \).
To formalize this, we propose the following optimization problem:
\small{
\begin{equation}
\mathcal{Q}^* = \underset{\mathcal{Q}}{\arg \min} \mathbb{E}_{\phi \sim \Phi} \left[ L_{\text{attack}}\left( M\left( I \odot \phi(\mathcal{Q}) \right) \right) +\lambda \cdot L_{\text{env}}\left( \mathcal{Q}, \mathcal{P} \right) \right], 
\end{equation}}
where $\mathcal{Q}$ is the adversarial patch to be optimized; $\phi(\cdot)$ represents the transformation applied to the patch $\mathcal{Q}$, simulating real-world environmental changes; $L_{\text{attack}}$ is the attack loss, encouraging the patch to deceive the detector by minimizing detection confidence; $\mathbb{E}_{\phi \sim \Phi}$ is the expectation over transformations $\Phi$; $L_{\text{env}}\left( \mathcal{Q}, \mathcal{P} \right)$ is the environmental consistency loss, which ensures that the patch blends naturally with the environment described by the prompt $\mathcal{P}$, e.g., ``forest'', ``desert'', etc; and $\lambda$ is a regularization term that balances the attack and environmental consistency losses.


By optimizing this joint objective, the resulting patch not only maximizes the success of the attack but also maintains environmental coherence, making it robust to real-world variations.

\subsection{Diffusion Model for Environment-Consistent Adversarial Patch Generation}
To generate adversarial patches that are both effective and environment-consistent, we propose a diffusion model-based approach. Unlike traditional pixel-space optimization, where adversarial patches are directly manipulated in the image space, our method operates in the \textbf{latent space} of the diffusion model, optimizing the latent variable \( z_T \) to generate the patch that consistent with the environment.

In the diffusion model, the denoising process progressively refines  a noisy latent variable \( z_T \) towards a clean latent variable \( z_0 \), which is then decoded into the final adversarial patch. The core of our approach is the \textbf{conditioning} of the diffusion process on the environmental context, which is encoded as \( \mathcal{C} \). The environmental context \( \mathcal{C} \) can be a textual prompt describing the target environment (e.g., ‘forest’, ‘urban’, or ‘desert’). This conditioning is performed via a \textbf{cross-attention mechanism}, where the latent representation \( z_T \) attends to the environmental features \( \mathcal{C} \), ensuring the generated patch aligns with the specific characteristics of the environment.

The optimization process in the latent space is formulated as:
\begin{align}
\underset{z_T}{\arg \min} \mathbb{E}_{\phi \sim \Phi} \Big[ & L_{\text{attack}}\left( M\left( I \odot \phi(\text{Decoder}(z_0)) \right) \right) \Big] \nonumber \\
& + \lambda \cdot L_{\text{env}}\left( \mathcal{Q}, \mathcal{P} \right) \Big]
\end{align}
where \( z_T \) is the noisy latent variable at the final timestep \( T \),  \( \mathcal{Q} \) is the adversarial patch generated from the latent variable \( z_0 \), \( M \) denotes the object detection model,  \( \phi(\text{Decoder}(z_0)) \) is the adversarial patch decoded from \( z_0 \), \( L_{\text{attack}} \) is the attack loss ensuring the patch’s effectiveness at evading detection, \( L_{\text{env}} \) is the environmental consistency loss ensuring the patch aligns with the environmental characteristics specified by the prompt \( \mathcal{P} \), and  \( \lambda \) is a regularization parameter balancing both losses. 

\begin{figure}[t]
  \centering
  \includegraphics[width=1.0\columnwidth]{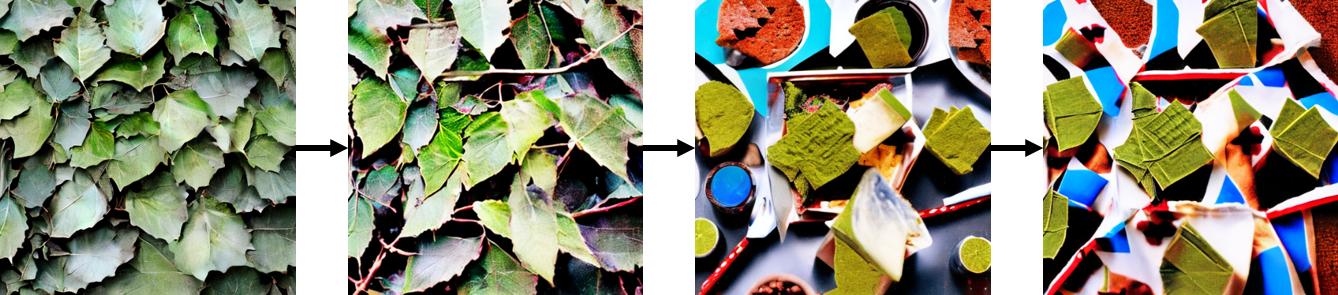}
  \caption{During the optimization of \( z_T \), without constraints, the adversarial patch gradually misaligns with  \( \mathcal{P} \), where \( \mathcal{P} \) is "a picture full of leaf-like green color".}
  \label{distortion}
\end{figure}

While the latent space optimization allows for robust adversarial patch generation, the optimization of \( z_T \) without additional constraints may result in patches that deviate from the desired environmental characteristics \( \mathcal{P} \) (see Fig.~\ref{distortion}). For example, the generated patch might fail to blend naturally with the background.


To address this, we introduce two regularization terms: \textbf{Prompt Alignment Loss} and \textbf{Latent Space Alignment Loss}, which guide the patch generation process to ensure both detection evasion and environmental consistency. The Prompt Alignment Loss ensures that the generated patch matches the characteristics described by the environmental prompt \( \mathcal{P} \). Latent Space Alignment Loss constrains the transformation process in the latent space, ensuring that the patch remains consistent with the environment during the denoising process. 
These two regularization terms are incorporated into the overall optimization framework, as described in the following sections. By introducing these constraints, we ensure that the adversarial patch is not only effective in deceiving object detectors but also robust and visually consistent across a variety of real-world environments.

\subsection{Prompt and Latent Space Alignment Losses}
\label{text_alignment_loss}
To generate adversarial patches that are both effective in evading detection and consistent with the environmental context $\mathcal{P}$, we introduce two key alignment losses during the optimization process: the \textbf{Prompt Alignment Loss} and the \textbf{Latent Space Alignment Loss}. These losses guide the optimization to ensure that the generated patches preserve both attack effectiveness and environmental consistency.

The first loss, \textit{Prompt Alignment Loss}, directly ensures that the generated adversarial patch aligns with the information embedded in the prompt $\mathcal{P}$. In diffusion models, the prompt $\mathcal{P}$ influences the latent variables through a cross-attention mechanism. Specifically, at each diffusion step, the model processes a latent variable $z_i$ to produce features $f_{i,j}$ at layer $j$. These features are then projected through learned matrices $W_Q$ and $W_K$ to obtain the query $Q$ and key $K$, which are used to compute the cross-attention map $\mathcal{A}_{i,j}$:
\begin{equation}
\mathcal{A}_{i,j} = \operatorname{softmax}\left(\frac{(f_{i,j} W_Q)(\mathcal{C} W_K)^T}{\sqrt{d}}\right)
\end{equation}

Here, $\mathcal{C}$ represents the text embeddings derived from the prompt $\mathcal{P}$, and the attention map $\mathcal{A}_{i,j}$ encodes the relevance of the features $f_{i,j}$ to the prompt. The text embeddings are also used to generate the context vector $V$ through a projection by matrix $W_V$, and the final attention output is computed by multiplying $\mathcal{A}_{i,j}$ with $V$.

It is important to note that, since \( V \) remains fixed across all diffusion steps, the diffusion model’s ability to embed \( \mathcal{P} \) is influenced solely by changes in \( \mathcal{A}_{i,j} \). Additionally, \( \mathcal{A}_{i,j} \) is affected only by changes in \( z_i \), as all other variables used in its computation remain fixed. As shown in Eq.~\ref{diffusion_pipeline}, any modification in \( z_T \) leads to changes in each intermediate latent variable \( z_i \), resulting in corresponding adjustments to \( \mathcal{A}_{i,j} \). If \( z_T \) deviates too much from its initial value during optimization, \( \mathcal{A}_{i,j} \) will also diverge, weakening the model’s ability to effectively embed \( \mathcal{P} \), as illustrated in Fig.~\ref{distortion}.

To prevent this misalignment, we track the initial cross-attention maps $\mathcal{A}_{i,j}^{\text{initial}}$ and align them with the updated maps during optimization using cosine similarity. The alignment loss for the cross-attention maps is thus defined as:
\begin{equation}
L_{\text{prompt}} = 1 - \frac{1}{N \cdot M} \sum_{i=1}^{N} \sum_{j=1}^{M} \frac{\mathcal{A}_{i,j} \cdot \mathcal{A}_{i,j}^{\text{initial}}}{\|\mathcal{A}_{i,j}\| \|\mathcal{A}_{i,j}^{\text{initial}}\|}
\end{equation}

This loss ensures the adversarial patch remains aligned with the environmental context represented by $\mathcal{P}$ throughout the optimization process.

To further ensure the generated adversarial patches remain faithful to $\mathcal{P}$, we additionally conduct alignment in the latent space. Specifically, the initial latent variable \( z_0^{\text{initial}} \), which has extensively interacted with the text throughout the entire \( T \)-step denoising process, encodes rich semantic information from \( \mathcal{P} \). Thus, we align the latent variable \( z_0 \) with its initial counterpart \( z_0^{\text{initial}} \) during the optimization to further enhances the consistency between the adversarial patch and \( \mathcal{P} \). The latent space alignment loss is defined as:

\begin{equation}
L_{\text{latent}} = 1 - e^{-(z_0 - z_0^{\text{initial}})^2}
\end{equation}

This loss helps preserve the consistency of the adversarial patch with the prompt's semantic content, further enhancing the alignment of the patch with the environment.

\begin{table}[t]
\centering 
\resizebox{\columnwidth}{!}{ 
\begin{tabular}{cccccc}
\toprule
\textbf{Detectors} & \textbf{Random} & \textbf{Gray} & \textbf{DAP} & \textbf{NAP} & \textbf{PG-ECAP} \\
\midrule
\textbf{Yolov2}         & 57.47 & 57.81 & 27.74 & 17.73 & \textbf{9.70} \\
\textbf{Yolov3}         & 78.80 & 79.29 & 42.19 & 47.73 & \textbf{35.12} \\
\textbf{Yolov4}         & 78.58 & 79.90 & \textbf{20.09} & 64.16 & 40.76 \\
\textbf{Yolov5}         & 79.18 & 79.95 & 10.26 & \textbf{5.95} & 16.00 \\
\textbf{Faster-rcnn}    & 70.40 & 67.16 & 54.08 & 42.47 & \textbf{35.26} \\
\textbf{DETR}           & 37.30 & 32.00 & \textbf{17.80} & 27.40 & 25.70 \\
\midrule
\textbf{Avg.}           & 66.96 & 66.02 & 28.69 & 34.24 & \textbf{27.09} \\
\bottomrule
\end{tabular}
}
\caption{The white-box attack performance on the INRIA dataset, reported as $mAP_{50}$ (lower is better). The first column is the white-box models.}
\label{vs_sota}
\end{table}

Finally, to generate adversarial patches that are both effective at evading detection and consistent with the environmental context $\mathcal{P}$, we combine the attack loss $L_{\text{attack}}$, the prompt alignment loss $L_{\text{prompt}}$, and the latent space alignment loss $L_{\text{latent}}$. The final optimization objective is a weighted sum of these losses, defined as:
\begin{equation}
\label{total_loss}
\underset{z_T}{\arg \min } L = \alpha L_{\text{attack}} + \beta L_{\text{prompt}} + \gamma L_{\text{latent}}
\end{equation}
where $\alpha$, $\beta$, and $\gamma$ are the weights assigned to each loss function. This comprehensive objective ensures that the generated adversarial patches effectively fool object detection models while maintaining consistency with the environmental features described by the prompt $\mathcal{P}$.

\begin{table}[t]
\centering 
\resizebox{1.0\columnwidth}{!}{
\begin{tabular}{ccccc}
\toprule
\textbf{Detectors} & \textbf{Yolov2}        & \textbf{Yolov3}         & \textbf{Yolov4}         & \textbf{Yolov5}         \\
\midrule
\textbf{Yolov2}    & \textbf{9.70} & 36.04          & 52.59          & 55.86          \\
\textbf{Yolov3}    & 48.60         & \textbf{35.12} & 52.78          & 56.81          \\
\textbf{Yolov4}    & 35.80         & 47.19          & \textbf{40.76} & 42.26          \\
\textbf{Yolov5}    & 32.17         & 56.43          & 49.3           & \textbf{16.00} \\
\bottomrule
\end{tabular}
}
\caption{The black-box attack performance of one-stage models on the INRIA dataset. The model in the first column is held out for the white-box model, while the remains are black-box models.}
\label{transfer}
\end{table}

\section{Experiments}
\label{sec
} In this section, we present the experimental results obtained from both the digital and physical worlds to demonstrate the effectiveness of our method.

\subsection{Experiment Settings}
\label{Experiment Settings}

\subsubsection{Datasets and Victim Models}
We conduct experiments using the widely adopted INRIA dataset~\cite{INRIA}, which consists of 614 training images and 288 test images captured in various environments. Each image is annotated with bounding boxes around pedestrians, allowing for effective training and evaluation of adversarial patches. We evaluate our method using a range of victim models, including single-stage models such as YOLOv2~\cite{yolov2}, YOLOv3~\cite{yolov3}, YOLOv4~\cite{yolov4}, and YOLOv5~\cite{yolov5}, as well as the two-stage model Faster R-CNN~\cite{faster_rcnn} and the transformer-based model DETR~\cite{ZhuSLLWD21}. All models are pretrained on the COCO dataset~\cite{coco}, and input images are resized to a resolution of $416 \times 416$. This diverse set of models allows us to evaluate the robustness of our adversarial patches across different detection architectures.

\begin{table*}[t]
\centering
\renewcommand{\arraystretch}{1.8}  
\begin{adjustbox}{max width=\textwidth}
\begin{tabular}{c|ccccccccc}
\toprule
\raisebox{4.0\height}{\textbf{Clothes}} & 
\multicolumn{1}{c}{\includegraphics[width=2.5cm]{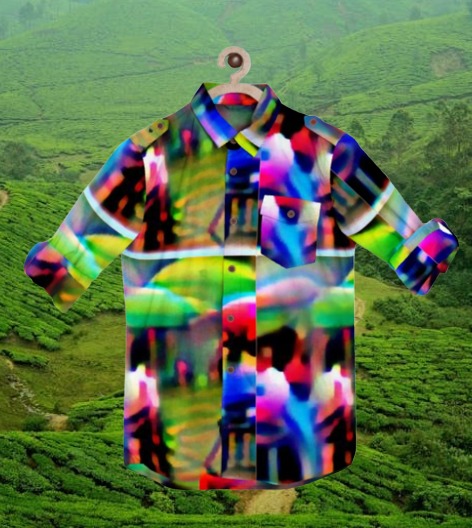}} & 
\multicolumn{1}{c}{\includegraphics[width=2.5cm]{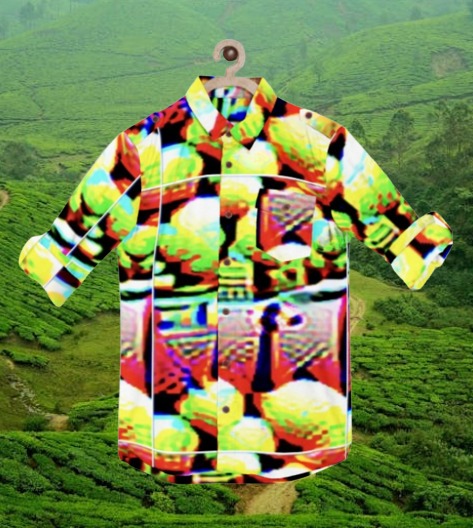}} & 
\multicolumn{1}{c}{\includegraphics[width=2.5cm]{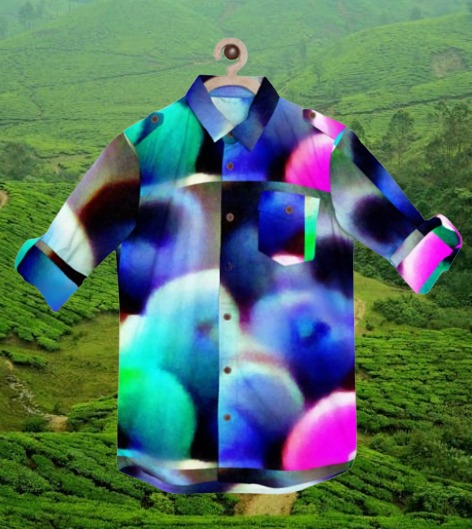}} & 
\multicolumn{1}{c}{\includegraphics[width=2.5cm]{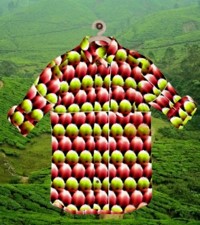}} &
\multicolumn{1}{c}{\includegraphics[width=2.5cm]{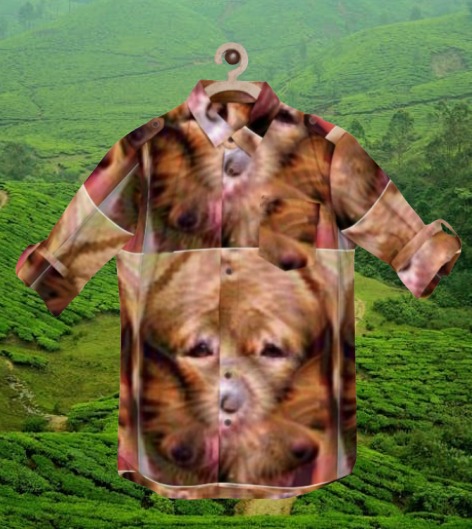}} &
\multicolumn{1}{c}{\includegraphics[width=2.5cm]{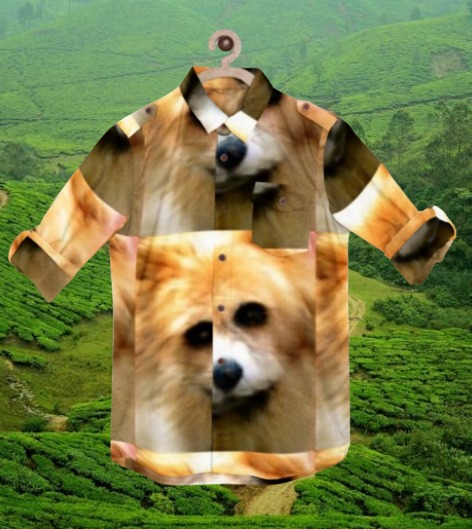}} &
\multicolumn{1}{c}{\includegraphics[width=2.5cm]{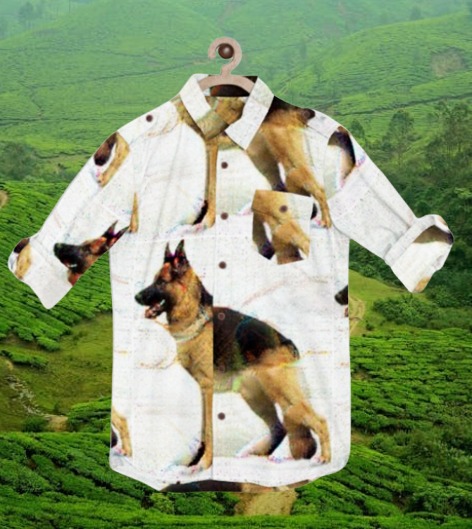}} &
\multicolumn{1}{c}{\includegraphics[width=2.5cm]{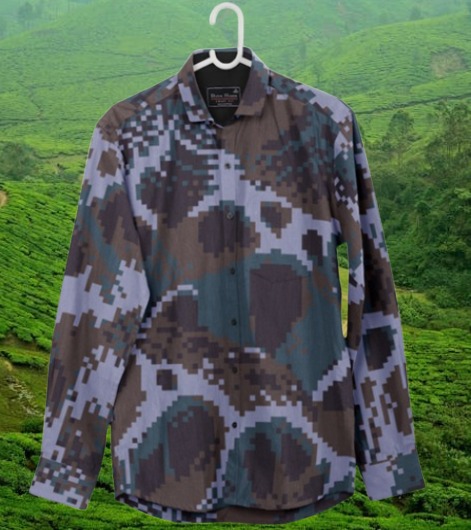}} &
\multicolumn{1}{c}{\includegraphics[width=2.5cm]{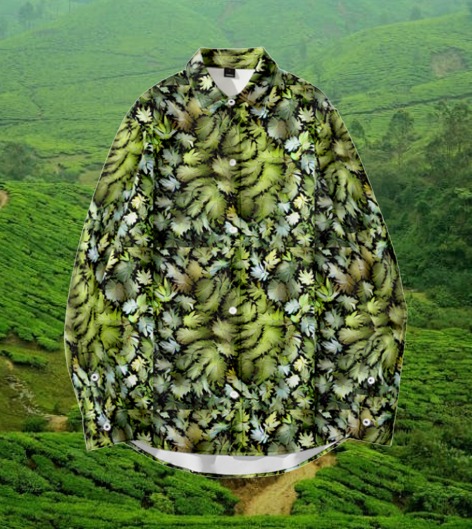}} \\
\midrule
\textbf{Score} & $2.78 \pm 1.64$ & $3.00 \pm 2.05$ & $1.91 \pm 1.08$ & $2.50 \pm 1.54$ & $1.78 \pm 0.83$ & $1.61 \pm 0.70$ & $1.44 \pm 0.73$ & $2.89 \pm 1.52$ & $5.56 \pm 1.67$\\
\midrule
\textbf{Source} &AdvYolo~\cite{thys2019fooling} &AdvT-shirt~\cite{xu2020adversarial} &T-SEA~\cite{huang2023t} &AdvTexture~\cite{hu2022adversarial} &UPC~\cite{huang2020universal} &NAP~\cite{hu2021naturalistic} &DAP~\cite{guesmi2024dap} &AdvCat~\cite{hu2023physically} &Ours \\
\bottomrule
\end{tabular}
\end{adjustbox}
\caption{Subjective evaluation is conducted on a 7-point Likert scale, ranging from 1 (completely inconsistent with the environment) to 7 (highly consistent with the environment). Our adversarial patch is generated by attacking Yolov5.}
\label{patch_consis}
\end{table*}

\begin{figure*}[t]
  \centering
  \includegraphics[width=1.0\textwidth]{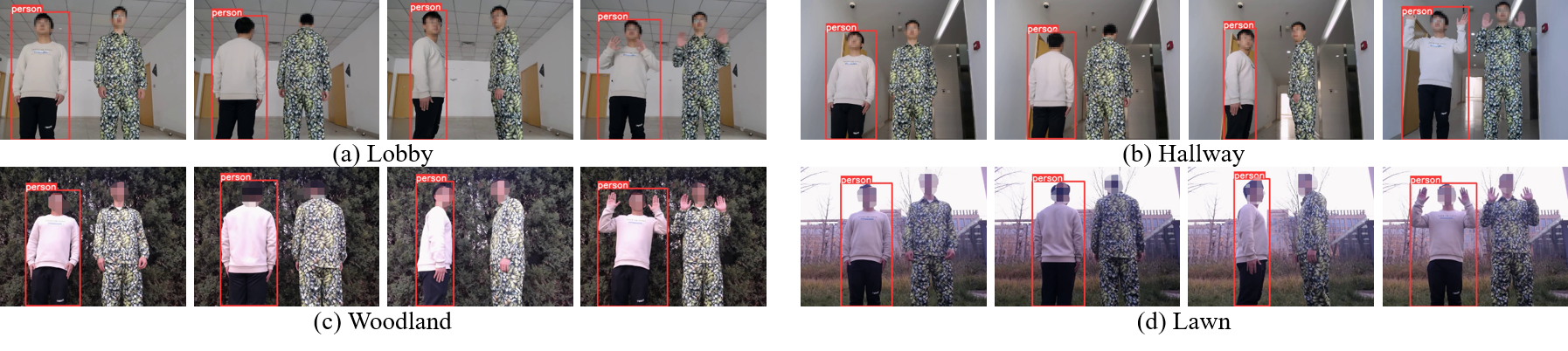}
  \caption{The detection results of four postures in four scenes. Our generated clothing can successfully evade detection in different scenes with different postures.}
  \label{Fig:physical}
\end{figure*}

\begin{table}[t]
\centering 
\resizebox{1.0\columnwidth}{!}{
\begin{tabular}{ccccc}
\toprule
\textbf{Scene} & \textbf{Lobby}        & \textbf{Hallway}         & \textbf{Woodland}         & \textbf{Lawn}         \\
\midrule
\textbf{mean ASR}    & 95.59 & 93.41          & 92.75          & 94.79          \\
\bottomrule
\end{tabular}
}
\caption{The mean ASR for each scene is calculated by averaging the ASR across four postures within that scene.}
\label{Tab:physical}
\end{table}

\subsubsection{Evaluation Metrics and Implementation Details}
Following previous works~\cite{hu2021naturalistic, hu2022adversarial}, we evaluate the effectiveness of our method in the digital world using the mean Average Precision at 50 IoU threshold ($mAP_{50}$). For physical-world experiments, we use the Attack Success Rate (ASR), defined as the ratio of successfully attacked images to the total number of images.

For image generation, we set the prompt $\mathcal{P}$ to "a picture full of leaf-like green colors" to represent a forest-like environment. We use DDIM~\cite{song2020denoising} as the sampler for Stable Diffusion 2~\cite{rombach2022high}, with denoising steps set to 7 and a guidance scale of 7.5. These parameters are chosen to balance image quality with computational efficiency. To optimize the latent vector $z_T$, we employ the Adam optimizer~\cite{kingma2014adam} with a learning rate of $5 \times 10^{-3}$ and train for 100 epochs. The weight factors $\alpha$, $\beta$, and $\gamma$ in Eq.~\ref{total_loss} are set to 1, 5, and 0.1, respectively, ensuring that the generated adversarial patches are both effective and visually natural.

\subsection{Attack Performance in the Digital World}
\label{Digital Attack}

\subsubsection{PG-ECAP vs. State-of-the-Art Methods}
To validate the effectiveness of our proposed method, we compare it with two state-of-the-art approaches in a white-box attack setting: NAP~\cite{hu2021naturalistic} and DAP~\cite{guesmi2024dap}. For benchmarking purposes, we also include two additional patch types: a randomly initialized patch and a gray patch, which we denote as "Random" and "Gray." Table~\ref{vs_sota} presents the results of the white-box attack on the INRIA dataset.

The results show that our proposed method achieves the best average attack performance with an $mAP_{50}$ of 27.09, which is 1.6 points lower than the next best result, DAP, with an $mAP_{50}$ of 28.69. This difference underscores the effectiveness of our method in evading detection. Additionally, our method demonstrates particular strength in attacking specific detectors. For example, PG-ECAP scores 9.70 when attacking YOLOv2, significantly outperforming other methods. Overall, these results highlight that PG-ECAP generates adversarial patches capable of effectively evading detection and achieving robust performance across various object detectors.

\subsubsection{Attacking Transferability Analysis} 
To further evaluate the effectiveness of our method, we perform a transferability analysis across a range of single-stage detectors, including Yolov2, Yolov3, Yolov4, and Yolov5. The results, which are summarized in Tab.~\ref{transfer}, demonstrate that our method exhibits strong transferability across different detectors, even though transferability was not explicitly considered into the optimization of $z_T$. For example, when Yolov4 is used as the white-box model, the average \( mAP_{50} \) on black-box models reaches 41.75, which is remarkably close to the \( mAP_{50} \) of 40.76 achieved when testing against Yolov4 itself. Future research can focus on exploring strategies to further improve the transferability of adversarial patches across different detectors.

\subsubsection{Subjective Evaluation of Patch Consistency}
\label{Subjective Evaluation}
To assess the environmental consistency of our adversarial patches compared to previous methods, we perform a subjective evaluation based on established procedures from prior work~\cite{hu2023physically}. For the background, we choose a green natural theme, with the specific background image sourced from Fig.~\ref{pipeline}. For T-shirt production, we print the adversarial patches generated by each method onto T-shirts. Specifically, to ensure a fair comparison, we use FAB3D\footnote{https://tri3d.in/} to tile 2D adversarial patches onto T-shirts, allowing us to compare them with methods that inherently generate 3D patches. Each T-shirt features at least one fully printed 2D adversarial patch on the front, with additional areas covered by tiling the same patch multiple times to ensure full surface coverage. For the evaluation, we use a 7-point Likert scale to assess the degree of consistency between the adversarial patches and the environment. A score of 1 indicates complete inconsistency, while a score of 7 represents perfect consistency. In total, 10 participants evaluate T-shirts generated by 9 different methods, scoring each T-shirt in a randomized order to eliminate bias. After the evaluation, we calculate the mean and standard deviation of the scores for each method, as summarized in Tab.~\ref{patch_consis}. The results show a clear advantage of our approach, which achieves the highest mean score of 5.56. This significantly outperforms the second-best method, which scores 3.00, with a notable difference of 2.56 points. This substantial gap highlights the superior effectiveness of our method in generating adversarial patches that align with the environment, demonstrating its ability to maintain environmental consistency compared to existing techniques.

\begin{figure}[t]
  \centering
  \includegraphics[width=1.0\columnwidth]{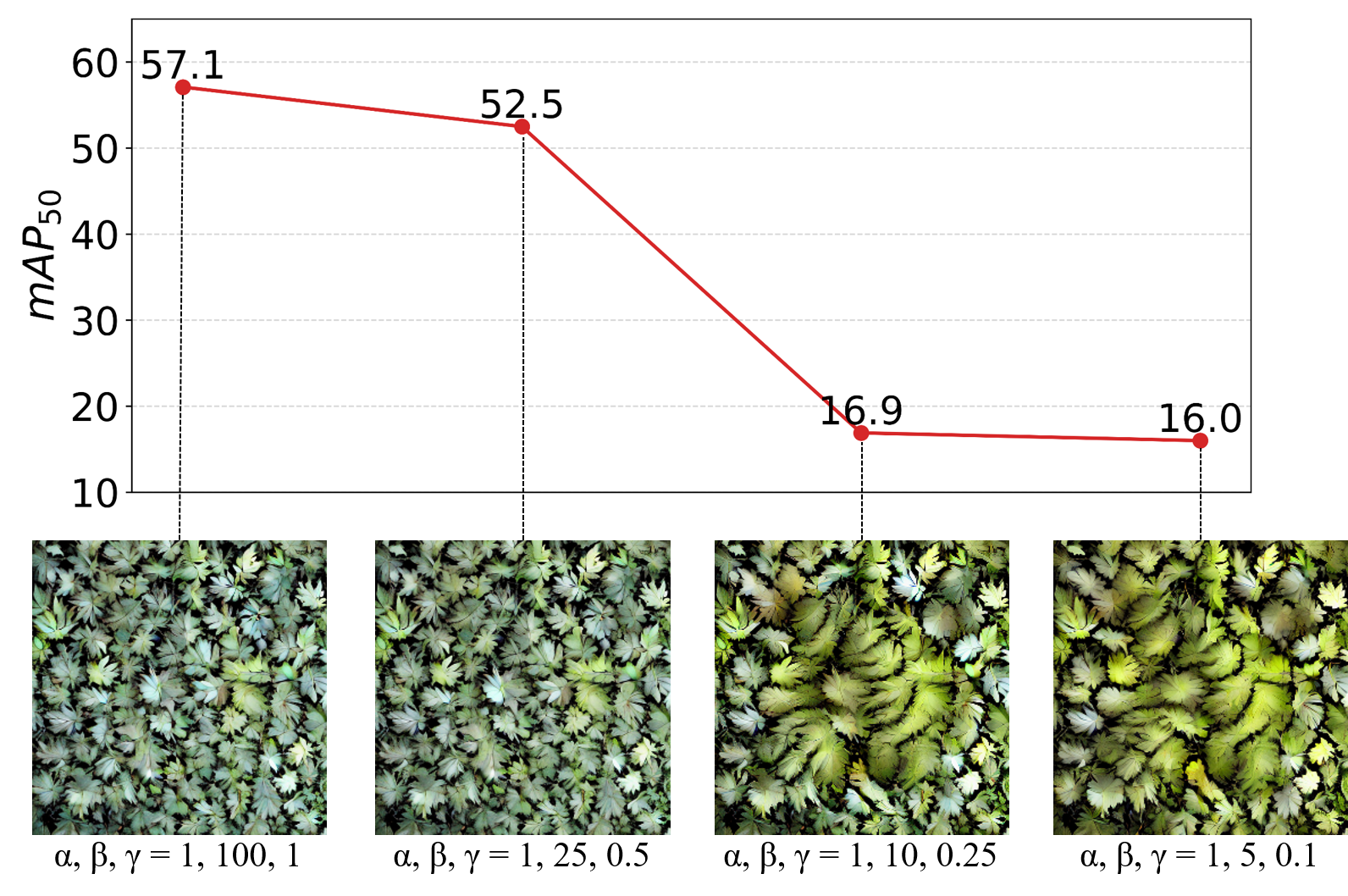}
  \caption{The attack performance and visualization of patches generated under different loss weight settings in Eq.~\ref{total_loss} are presented. We simultaneously decrease \( \beta \) and \( \gamma \) to find the optimal weight configuration, with the specific weight setting noted below each patch's visualization. All patches are generated and evaluated on Yolov5.}
  \label{ablation_wc}
\vspace{-0.3cm}
\end{figure}
\subsection{Attack Performance in the Physical World}
\label{Physical Attack}
To verify the feasibility of our PG-ECAP in the physical world, we tile the generated patch, designed to attack Yolov5, onto a long-sleeved T-shirt using FAB3D and produce a physical version. To thoroughly validate the effectiveness of our T-shirt, we test it in four scenes: two indoor settings (Lobby and Hallway) and two outdoor settings (Woodland and Lawn), with participants in four postures (front, back, side, and wave). For comparison, we also include a participant wearing regular clothing. To assess the attack success rate (ASR), we record each posture for 30 seconds, capturing 10 frames per second, resulting in a total of 300 frames. We then calculate the ASR as the ratio of frames that evade detection to the total number of frames.

We average the ASR across the four postures to obtain the mean ASR for each scene. The detection results and mean ASR are shown in Fig.~\ref{Fig:physical} and Tab.~\ref{Tab:physical}, respectively. The results indicate that our generated T-shirt achieves a high ASR across various scenes, with a average ASR of 94.14\% across the four scenes. Notably, despite the full patch being applied only to the front and back of the T-shirt, the side posture also evades detection, suggesting that the local pattern of our patch may also disrupt the detector's performance.

\subsection{Ablation Study}
\label{Ablation Study}

\subsubsection{Different Loss Weight Settings}
To evaluate the effectiveness of our proposed alignment losses and find the optimal configuration for the loss weights in Eq.~\ref{total_loss}, we conduct an ablation study by simultaneously reducing \( \beta \) and \(\gamma \). This helps us better understand how these parameters interact and their impact on patch generation. The results, shown in Fig.~\ref{ablation_wc}, indicate that higher values of \( \beta \) and \( \gamma \) produce patches with a more natural appearance, but at the cost of relatively lower attack performance. In contrast, reducing \( \beta \) and \( \gamma \) significantly improves attack performance while maintaining an acceptable level of visual quality. This shows that our alignment losses effectively preserve the natural appearance of the patches. When \( \beta \) and \( \gamma \) are set to 10 and 0.25, and 5 and 0.1, respectively, both attack effectiveness and visual quality stabilize, resulting in a balanced outcome. Based on these findings, we select \( \alpha = 1 \), \( \beta = 5 \), and \( \gamma = 0.1 \) as the optimal weight configuration. 

\subsubsection{Adversarial Patches for Different Environments}
To assess the adaptability of our proposed method across various environments, we conduct an ablation study by changing the prompt $\mathcal{P}$ to guide the patch generation process. Specifically, we use two additional prompts: 'a picture of a desert grid style' and 'a picture of an ocean-like pattern.' These prompts are then used to generate adversarial patches, which are trained and evaluated on YOLOv5.

After training, we use FAB3D to tile the patch onto a T-shirt, as described in Sec.~\ref{Subjective Evaluation}. The resulting T-shirt and its attack performance in the digital world are shown in Tab.~\ref{ablation_de}. From the figure, it is clear that the generated patches match well with their corresponding prompts $\mathcal{P}$. Furthermore, the attack performances are satisfactory. The original $mAP_{50}$ of the INRIA dataset on Yolov5 is 91.5, while the average $mAP_{50}$ of the two newly generated patches is 19.9. These results demonstrate the adaptability of our method, showing its potential to perform well across different environments.

\begin{table}[t]
\centering
\renewcommand{\arraystretch}{1.0}
\begin{adjustbox}{max width=\textwidth}
\begin{tabular}{c|cc}
\toprule
\raisebox{4.0\height}{Clothes} & 
\multicolumn{1}{c}{\includegraphics[width=2.5cm]{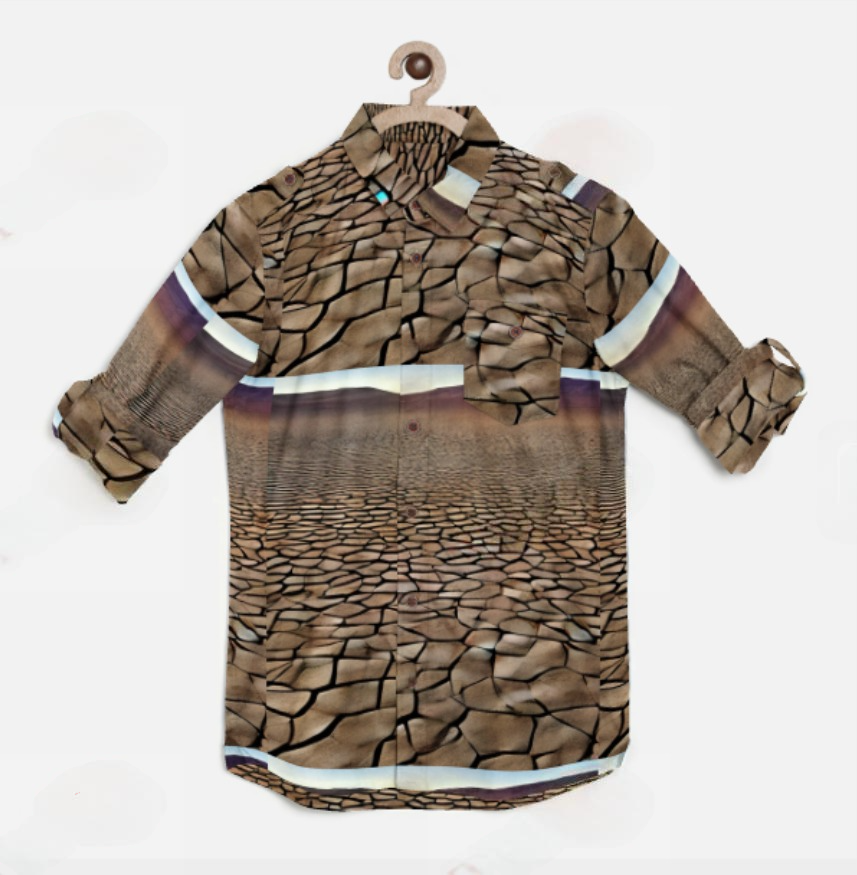}} &
\multicolumn{1}{c}{\includegraphics[width=2.5cm]{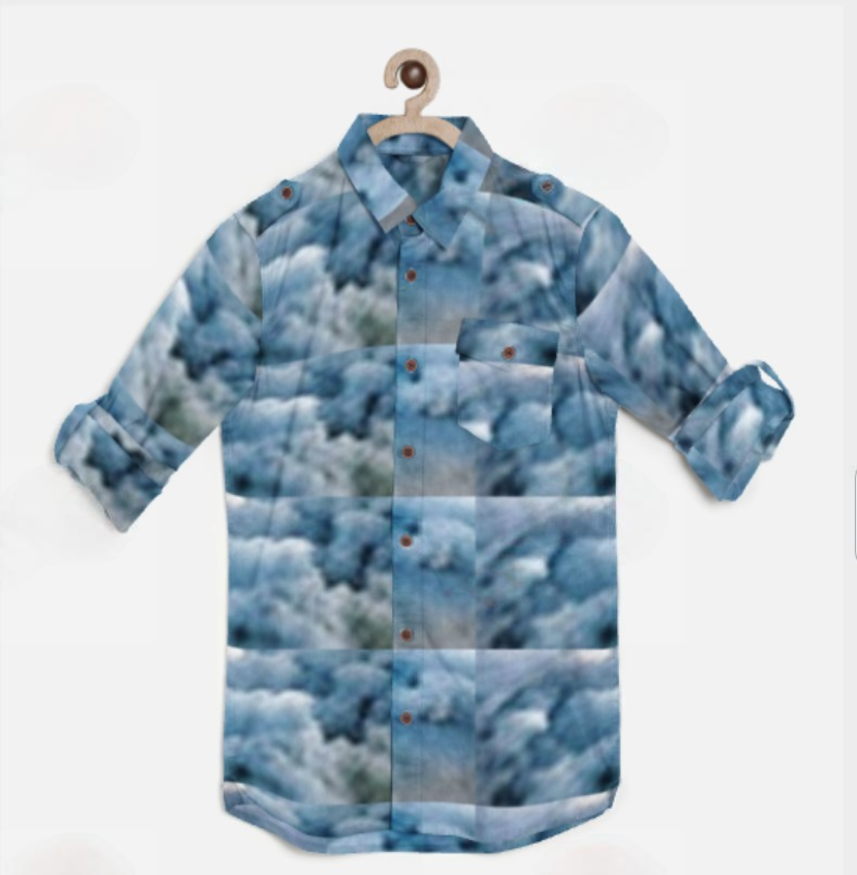}} \\
\midrule
$mAP_{50}$ & 21.7 & 18.1\\
\bottomrule
\end{tabular}
\end{adjustbox}
\caption{Our adversarial clothing generated for different environments. We use two additional prompts: "a picture of a desert grid style" and "a picture of an ocean-like pattern" to guide the patch generation process.}
\label{ablation_de}
\end{table}

\section{Conclusion}
\label{sec:conclusion}

In this paper, we present PG-ECAP, a novel method for generating environmentally consistent adversarial patches using diffusion models and two types of alignment losses. Unlike existing approaches that often prioritize either attack performance or naturalness at the expense of the other—leading to conspicuous and environmentally inconsistent patches—PG-ECAP strikes a balance by producing patches that seamlessly blend with their environment. By leveraging the text-to-image capabilities of diffusion models, our method uses environment-matching prompts to guide the patch generation process, ensuring that the generated patches appear natural. The Prompt Alignment Loss and Latent Space Alignment Loss further guarantee that the patches are aligned with the given prompts. Experimental results show that PG-ECAP outperforms state-of-the-art methods, achieving superior attack success rates while maintaining high environmental consistency across a variety of scenarios.



{
    \small
    \bibliographystyle{ieeenat_fullname}
    \bibliography{main}
}

\end{document}